%% file: anonymous-submission-latex-2025.tex
\documentclass[letterpaper]{article} % DO NOT CHANGE THIS
\usepackage{aaai25}  % DO NOT CHANGE THIS
\usepackage{times}  % DO NOT CHANGE THIS
\usepackage{helvet}  % DO NOT CHANGE THIS
\usepackage{courier}  % DO NOT CHANGE THIS
\usepackage[hyphens]{url}  % DO NOT CHANGE THIS
\usepackage{graphicx} % DO NOT CHANGE THIS
\usepackage{natbib}  % DO NOT CHANGE THIS AND DO NOT ADD ANY OPTIONS TO IT
\usepackage{caption} % DO NOT CHANGE THIS AND DO NOT ADD ANY OPTIONS TO IT
\usepackage{algorithm}
\usepackage{algorithmic}
\usepackage{inconsolata}
\usepackage{amsmath,amssymb,amsfonts}
\usepackage{algorithmic}
\usepackage{graphicx}
\usepackage{subfigure}
\usepackage{textcomp}
\usepackage{xcolor}
\newcommand{\answerYes}[1]{\textcolor{red}{Yes}} 
\newcommand{\answerNo}[1]{\textcolor{red}{No}} 
\newcommand{\answerNA}[1]{\textcolor{red}{NA}}

\usepackage{enumitem}
\usepackage{bm}
\usepackage{multirow}
\usepackage{booktabs}
\usepackage{graphics}
\usepackage{keyval}
\usepackage{trig}
\usepackage{threeparttable}
\usepackage{marvosym}

\usepackage{newtxmath}
\usepackage{tabularx}

\usepackage{newfloat}
\usepackage{listings}
\DeclareCaptionStyle{ruled}{labelfont=normalfont,labelsep=colon,strut=off} % DO NOT CHANGE THIS
\floatstyle{ruled}
\newfloat{listing}{tb}{lst}{}
\floatname{listing}{Listing}
\title{Importance Weighting Can Help Large Language Models Self-Improve}
\author{
    Chunyang Jiang\textsuperscript{\rm 1},
    Chi-Min Chan\textsuperscript{\rm 1},
    Wei Xue\textsuperscript{\rm 1 \Letter},
    Qifeng Liu\textsuperscript{\rm 1},
    Yike Guo\textsuperscript{\rm 1 \Letter}
}
\affiliations{
    \textsuperscript{\rm 1}Hong Kong University of Science and Technology\\
    \{cjiangaq, cchanbc\}@connect.ust.hk, \{weixue, liuqifeng, yikeguo\}@ust.hk
}

\usepackage{bibentry}
\begin{document}

\maketitle
\begin{abstract}
Large language models (LLMs) have shown remarkable capability in numerous tasks and applications. However, fine-tuning LLMs using high-quality datasets under external supervision remains prohibitively expensive. In response, LLM self-improvement approaches have been vibrantly developed recently. The typical paradigm of LLM self-improvement involves training LLM on self-generated data, part of which may be detrimental and should be filtered out due to the unstable data quality. While current works primarily employs filtering strategies based on answer correctness, in this paper, we demonstrate that filtering out correct but with high \textit{distribution shift extent} (DSE) samples could also benefit the results of self-improvement. Given that the actual sample distribution is usually inaccessible, we propose a new metric called \textit{DS weight} to approximate DSE, inspired by the \textit{Importance Weighting} methods. Consequently, we integrate DS weight with self-consistency to comprehensively filter the self-generated samples and fine-tune the language model. Experiments show that with only a tiny valid set (up to 5\% size of the training set) to compute DS weight, our approach can notably promote the reasoning ability of current LLM self-improvement methods. The resulting performance is on par with methods that rely on external supervision from pre-trained reward models.
\end{abstract}
\input{sections/intro}
\input{sections/related}
\input{sections/method}
\input{sections/experiment}
\input{sections/conclusion}
\section{Acknowledgement}
This research was supported by Theme-based Research Scheme (T45-205/21-N)
from Hong Kong RGC, and Generative AI Research and Development Centre
from InnoHK. The corresponding authors are Wei Xue and Yike Guo.
% \input{sections/Limitation}

\bibliography{aaai25}

\newpage
\input{sections/appendix}

\end{document}

%% file: sections/intro.tex
\section{Introduction}
Recently, Large Language Models (LLMs) have made impressive achievements on a large amount of NLP tasks and applications~\citep{CAMEL,GPT-4,leandojo, StarCoder}. Moreover, new capabilities emerge in LLMs with the model size scaled to hundreds of billions of parameters, especially the general reasoning capabilities~\citep{step-by-step}. Relevant techniques like in-context few-shot learning~\citep{in-context}, Chain-of-Thought prompting~\citep{CoT}, and self-consistency~\citep{self-consistency} were further proposed to get better performance.

Despite the remarkable capabilities of LLMs pre-trained on the large corpus, fundamentally improving the model's performance still necessitates fine-tuning on a great amount of high-quality supervised data~\citep{LMSI}, which is usually costly. To alleviate this problem, many works are committed to investigating the \textit{self-improvement} ability of LLMs~\citep{reflextion, self-refine, small-rewrite}. Among them, fine-tuning the LLM on self-generated data appears as one of the most promising way~\citep{rest, LMSI, self-instruct, WizardLM, Self-alignment}. This formula typically includes generating reasoning thoughts and answers on unsupervised datasets, filtering data, and fine-tuning models on the self-generated data~\citep{LMSI}. It is regarded as an attractive approach for LLMs to self-supervise by utilizing unlabeled data without external supervision.

% Self-improvement methods leverage the inherent knowledge of a pre-trained language model to provide feedbacks~\citep{reflextion}, make refinement~\citep{self-refine, small-rewrite}, or generate self-training data~\citep{LMSI, self-instruct, rest} in order to promote model performance without external supervision.

% Fine-tuning the LLM on self-generated data appears as one of the most promising way among self-improvement methods. This formula typically includes generating reasoning thoughts and answers on unsupervised datasets, filtering data, and fine-tuning models on these self-generated data~\citep{rest, LMSI, self-instruct, WizardLM}. 
% The most attractive feature of this kind of method is that it serves as an approach for LLMs to self-evolve by utilizing unsupervised data, which is much easier to obtain than high quality labeled data.

The primary challenge of utilizing self-generated data is the variability in data quality. While high-quality samples can enhance the model's reasoning abilities, there are low-quality samples that may detrimentally affect performance~\citep{MoT}. For example, an incorrectly generated answer could mislead the model. Therefore, a good filtering strategy is decisive for effective self-improvement. Many approaches have been proposed to address this issue.
Inspired by Self-Consistency~\citep{self-consistency}, LMSI~\citep{LMSI} adopts majority voting to select the most \textit{consistent} answer, under the assumption that consistency is positively related to the correctness. MoT~\citep{MoT} further introduces uncertainty to the filtering strategy, by utilizing entropy to exclude high-uncertainty data points. Self-Alignment~\citep{Self-alignment} demonstrates that prompting the LLM to self-filter is also feasible.
% Besides, there are works utilizing external pre-trained LLMs as the reward model to score the generated data, such as GENIE~\citep{GENIE} and ReST~\citep{rest}. Compared to LMSI, they introduced knowledge from external models and need more computational resources.

However, present methods mostly emphasize assessing the correctness of generated samples, yet ignore the distribution shift problem. Specifically, the distribution of the LLM self-generated data may differ from that of real-world data, and fine-tuning models on samples with high \textit{distribution shift extent} (DSE) may defect the resulting performance~\citep{moddel-collapse}. In this paper, we demonstrate that 
% the correlation between the correctness and DSE of a generated sample is relatively weak. In other words, 
even self-generated samples with correct answers can possess high DSE, potentially degrading model performance. Consequently, filtering out high DSE samples is essential to further promote the efficacy of LLM self-improvement.%不是很详细

To exclude samples with high DSE, the primary question is how to estimate the DSE, since the actual data distribution is usually inaccessible. We note \textit{Importance Weighting} (IW)~\citep{importance-weighting} as a well-known approach to address the traditional distribution shift problem~\citep{DS}, where the key idea is deriving importance weights based on the distribution ratio between test and training data, and using it to rebuild an unbiased training loss. IW usually contains two steps: \textit{weight estimation} computes test-over-training density ratio and \textit{weighted classification} utilizes the ratio to weight each data point and train the model~\citep{DIW}.

Inspired by IW, we propose \textit{Distribution Shift Weight} (DS weight) as a new metric to measure the DSE of self-generated samples. Based on this, we build an LLM self-improvement framework that incorporates both the correctness and DSE in its filtering strategy. Specifically, given a question-only dataset, we first let a pre-trained LLM generate multiple reasoning thoughts as well as answers. Then we create a tiny valid set comprising a few human-written demonstrations. With the pre-trained LLM and valid set, we leverages a simple approximation for importance weights to compute DS weight, as a measure of DSE, for each training data point. We subsequently combine the results from majority voting (for correctness) and DS weight (for DSE) to filter the dataset and fine-tune the LLM. We denote our framework as \textbf{I}mportance \textbf{W}eighting-based \textbf{S}elf-\textbf{I}mprovement (\textbf{IWSI}). Experiments show that the performance of IWSI largely surpasses baseline self-improvement methods and rivals the enhancements achieved with supervision from the pre-trained reward model.

Our contributions are threefold: (1) We propose a metric called DS weight to approximate the DSE of LLM self-generated data, with help from a tiny valid set. (2) Leveraging DS weight, we build a novel self-improvement framework called IWSI where the filtering strategy considers both the answer correctness and DSE. (3) We empirically examine the effectiveness of our proposed method, analyze the impact of high DSE samples on LLM self-improvement, and explore how DS weight interacts with other filtering criteria.
% verify the effectiveness of our proposed methods, discuss the influence of data with high DSE in LLM self-improvement, and investigate its relationship with other filtering factors.

%% file: sections/related.tex
\section{Related Work}
\input{figures/overview}
\subsection{LLM Self-Improvement}
Fundamentally improving LLMs' reasoning ability essentially requires fine-tuning on a large amount of high-quality supervised data. However, this methodology faces the threat that the stock of high-quality language data will be exhausted in some day~\citep{out-of-data}. Self-improvement emerges as a promising approach to utilize the inherent knowledge to make supervision for self-training LLMs. While LLMs can easily generate extensive data, the data quality is not always guaranteed~\citep{Hallucination} and training on unfiltered data may even cause performance degradation~\citep{recursion-curse}. Therefore, an essential requirement in LLM self-improvement is data filtering.

Pioneering works~\citep{self-instruct, cons-AI, WizardLM} use language models to generate diverse types of data such as feedback, instructions, and questions. They filter data by heuristic rules as well as manual inspection, which is challenging and costly. LMSI~\citep{LMSI} proposed a framework including generating data for a question-only dataset and using the majority voting (self-consistency)~\citep{self-consistency} to select the most consistent answers, which is empirically proven to be effective among various tasks. LMSI also demonstrates that the answer correctness is positively relevant to self-consistency. Along with this work, MoT~\citep{MoT} proposes further filtering the consistent answers by entropy, which measures the answer uncertainty. Self-Alignment~\citep{Self-alignment} shows it is feasible to prompt the LLM self-filtering the generated data. To comprehensively evaluate the generated data, some works use external pre-trained LMs as the reward model to score the generated data, such as GENIE~\citep{GENIE} and ReST~\citep{rest}. With external supervision from the reward model, their filtering strategies are typically more considered.

\subsection{Importance Weighting}
Importance weighting (IW) is a primary approach to mitigate the influence of distribution shift problem~\citep{DS}. The typical IW process includes two steps: weight estimation and weighted classification. Weight estimation approximates the importance weights, which are subsequently used in the weighted classification stage to build a weighted training loss~\citep{GIW}. 
% serves as the importance weight to weight the training loss function in weighted classification stage.

Traditional IW methods mainly estimate the importance weights by assessing the matching between training and test distribution in different ways, such as maximum mean discrepancy in a reproducing kernel Hilbert space~\citep{KMM}, KL divergence~\citep{Imprtance-estimation}, and squared loss~\citep{LSIF}. While these methods work well in linear models, their performances degrade largely in deep learning scenarios~\citep{DIW}. To overcome this, DIW~\citep{DIW} proposes an end-to-end dynamic solution, which uses a deep network to predict the importance weights, and repeats weight estimation and weighted classification stages to iteratively converge on the optimal solution.

In this paper, we use some lemmas and empirical results in DIW to build the DS weight for estimating the DSE of self-generated data.

%% file: figures/overview.tex
\begin{figure*}[t]
  \centering
  \label{fig:overview}
  \includegraphics[width=0.85\textwidth]{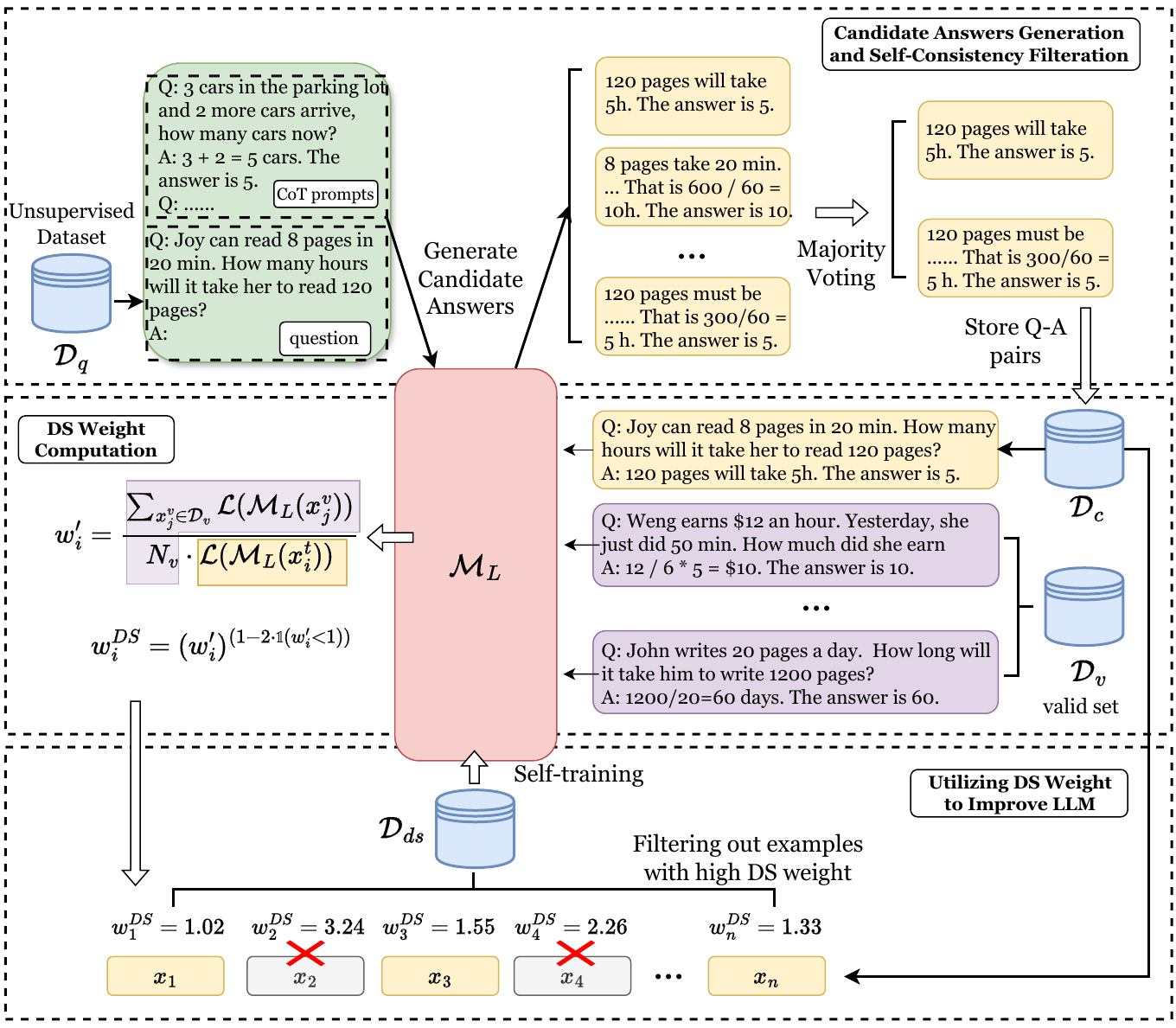}
  \caption{The overview of IWSI. Given the unsupervised dataset $\mathcal{D}_q$, the pre-trained LLM $\mathcal{M}_L$ is first used to generate multiple candidate answers as well as the reasoning thoughts, prompted by CoT examples. Then IWSI uses majority voting to select the most consistent answer and corresponding thoughts, stored in $\mathcal{D}_c$. With the help of $\mathcal{D}_v$, IWSI calculates DS weight for each data point in $\mathcal{D}_c$. IWSI filters $\mathcal{D}_c$ into $\mathcal{D}_{ds}$ by keeping samples with the $k\%$-lowest DS weight and lastly self-trains $\mathcal{M}_L$.}
  \label{fig:overview}
\end{figure*}

%% file: sections/method.tex
\section{Methodology}
Fig.~\ref{fig:overview} shows the overview of IWSI. Given an unsupervised (question-only) dataset $\mathcal{D}_q$, we first use the pre-trained LLM $\mathcal{M}_L$ to generate multiple candidate answers as well as the reasoning thoughts for each question, using CoT prompts~\citep{CoT}. 
% Consequently, we get the unfiltered self-generated dataset $\mathcal{D}$. 
Following LMSI~\citep{LMSI}, we adopt the majority voting to keep the most \textit{consistent} answer and corresponding thoughts for each question, resulting in the consistency-filtered dataset $\mathcal{D}_c$. Then we calculate \textit{DS weight} for every data point in $\mathcal{D}_c$, with the help of a tiny valid set $\mathcal{D}_v$. Lastly, we filter $\mathcal{D}_c$ into $\mathcal{D}_{ds}$ utilizing the DS weight and fine-tune the model $\mathcal{M}_L$. The following sections elaborate on different components of IWSI.

\subsection{Candidate Answers Generation and Self-Consistency Filtration}
In this stage, we let the pre-trained LLM $\mathcal{M}_L$ generate candidate answers as well as reasoning thoughts for an unsupervised dataset $\mathcal{D}_q$ which only contains unlabeled questions. Given a question $q_i \in \mathcal{D}_q$, we concatenate Few-Shot-CoT~\citep{CoT} prompts with $q_i$ to form the input text $x_i$. With temperature $T>0$, we let $\mathcal{M}_L$ sample $m$ candidate answers $[a_{i_1},a_{i_2}, \dots, a_{i_m}]$ and their reasoning thoughts $[r_{i_1},r_{i_2}, \dots, r_{i_m}]$. 
% The candidate answers $[a_{i_1},a_{i_2}, \dots, a_{i_m}]$ are derived by parsing corresponding reasoning paths with the pattern that the answer is just behind \textit{"The answer is"}. 
Then we select the most \textit{consistent} answer $\hat{a_i}$ by majority voting~\citep{self-consistency}, 
$\hat{a_i} = \arg\max_{a_{i_j}} \sum_{k=1}^m \vmathbb{1}(a_{i_j}=a_{i_k})$, and keep the corresponding reasoning thoughts $R_i = \{r_{i_j}|a_{i_j}=\hat{a_i}, 1 \leq j \leq m\}$. By repeating over each question in $\mathcal{D}_q$, the consistency-filtered dataset $\mathcal{D}_c$ is built.

\subsection{DS Weight Computation}
To elaborate \textit{DS Weight}, we first introduce some important preliminaries in the distribution shift problem and importance weighting methods.

\textit{Distribution shift} problem denotes that the training data and test data are drawn from two different distributions $p_{train}$ and $p_{test}$, and $p_{train} \neq p_{test}$~\citep{DS}. A common assumption for distribution shift is that there exists a function $w^*(x)$, holding that:
\begin{equation}
    \label{IW_equ}
    \vmathbb{E}_{p_{test}(x)}[f(x)] = \vmathbb{E}_{p_{train}(x)}[w^*(x)\cdot f(x)]
\end{equation}
for any function $f$ of $x$~\citep{DIW}. Based on Eq.~\ref{IW_equ}, importance weighting methods~\citep{importance-weighting, Imprtance-estimation} deal with distribution shift in two steps: \textit{weight estimation} finds a proper solution for $w^*(x)$; \textit{weighted classification} trains the model with a weighted loss derived by substituting $f$ in Eq.~\ref{IW_equ} with the target loss function.

Obviously, it plays a decisive role in importance weighting that finding the appropriate importance weights $\mathcal{W} = \{w_i\}^{N_{t}}$, to approximate $w^*(x)$ in Eq.~\ref{IW_equ}. To simplify the question, DIW~\citep{DIW} provides an empirical surrogate goal with the help of a valid set:
\begin{equation}
    \label{equ:DIW}
    \frac{1}{N_v} \sum_{j=1}^{N_v} \mathcal{L}(\mathcal{M}(x^v_j)) \approx \frac{1}{N_{t}} \sum_{i=1}^{N_{t}} w_i \cdot \mathcal{L}(\mathcal{M}(x^t_i)).
\end{equation}
Here $N_v$, $N_t$, $x^v$, and $x^t$ indicate the size of the valid set, the size of the training set, data in the valid set, and data in the training set. $\mathcal{M}$ is the training model and $\mathcal{L}$ represents the training loss.

While in DIW, Eq.~\ref{equ:DIW} is used as a goal to train a deep model that predicts the desired $\mathcal{W}$, we use Eq.~\ref{equ:DIW} to design a naive measurement for the distribution shift extent between training samples and valid set. Our intuition is that when the training data distribution is identical to the valid data distribution, $w_i \equiv 1$ would be a proper solution to Eq.~\ref{equ:DIW}. Conversely, the larger the actual $w_i$ differs from $1$, the more different the training distribution and valid distribution are. 

Based on this idea, we first design a naive estimation $w_i^\prime$ for $x_i^t$ by regarding $N_t$ as $1$:
\begin{equation}
\label{equ:weight-est}
    w_i^{\prime} = \frac{\sum_{x_j^v \in \mathcal{D}_v} \mathcal{L}(\mathcal{M}_L(x^v_j))}{N_v \cdot \mathcal{L}(\mathcal{M}_L(x^t_i))}
\end{equation}
where $\mathcal{M}_L$ is the pre-trained LLM, $\mathcal{L}$ denotes the \textit{sft} loss~\citep{GPT3}, $\mathcal{D}_v$ is a tiny valid set and $x_i^t$ is a self-generated training data point. Here we notice that the value range of $w_i^\prime$ is $(0,+\infty)$ while the ideal value is $1$, which creates asymmetry between the two deviation directions (lower than $1$ and greater than $1$) and makes filtering inconvenient. Therefore, to establish symmetry for both shift directions, we define \textit{DS weight} $w_i^{DS}$ as:
\begin{equation}
\label{equ:DS-weight}
    w_i^{DS}=
       \begin{cases}
       w_i^{\prime} & \mbox{if } w_i^{\prime}\geq1\\
       \frac{1}{w_i^{\prime}} & \mbox{if }w_i^{\prime}<1
       \end{cases}
\end{equation}

\subsection{Utilizing DS Weight to Improve LLM}
\label{sec:variants}
With DS weight approximating DSE, we are able to further filter the self-generated data in $\mathcal{D}_c$, excluding data points that possibly possess higher DSE.
% With DS weight approximating DSE, we propose two methods to improve the LLM. One is using DS weight to further filter the self-generated data in $\mathcal{D}_c$, followed by fine-tuning the model using these refined samples. The other directly employs Eq.~\ref{equ:weight-est} to calculate a weighted loss, analogous to the standard weighted classification procedure. We call these two variants \textit{IWSI-F} and \textit{IWSI-W} respectively.

% We first introduce IWSI-F, which further filters out data points that have high DSE. 
First, all data points are ranked with respect to their DS weight $w_i^{DS}$, and the $k$-percentile $\sigma_{k\%}$ is selected, \textit{s.t.}
\begin{equation}
\label{equ:sigma}
    \frac{\sum_{i}^{|\mathcal{D}_c|} \vmathbb{1}(w_i^{DS} \leq \sigma_{k\%})}{|\mathcal{D}_c|} = k\%
\end{equation}
where $|\cdot|$ denotes the set size and $w_i^{DS}$ is the corresponding DS weight of sample $x_i$. As a result, only samples whose $w_i^{DS} \leq \sigma_{k\%}$ are kept to train the model $\mathcal{M}_L$. The training loss can be written as:
\begin{equation}
    \label{IWSI-F}
    \mathcal{L}_{F} = \frac{1}{|\mathcal{D}_c|\cdot k\%}\sum_{x_i}^{\mathcal{D}_c}\vmathbb{1}_{k\%}(x_i)\cdot \mathcal{L}(\mathcal{M}_L(x_i))
\end{equation}
where $\vmathbb{1}_{k\%}(x_i)$ equals to $\vmathbb{1}(w_i^{DS} \leq \sigma_{k\%})$ and $\mathcal{L}$ represents the \textit{sft} loss.

Another natural way to utilize DS weight is directly employing Eq.~\ref{equ:weight-est} to calculate a weighted loss, which is more analogous to the standard IW procedure. We also implement this variant in our work and denote it as IWSI-w. The weighted loss is:
\begin{equation}
    \label{IWSI-W}
    \mathcal{L}_{W} = \frac{1}{|\mathcal{D}_c|}\sum_{x_i}^{\mathcal{D}_c} Clip( w_i^{\prime},C)\cdot \mathcal{L}(\mathcal{M}_L(x_i))
\end{equation}
where $C$ is a constant. We clip $w_i^{\prime}$ to $(0, C]$ for stabilizing the training process.

However, we found that IWSI-w is much less effective than IWSI. We believe this is mainly attributed to the inadequacy of Eq.~\ref{equ:weight-est}. Empirical results and details are discussed in the experiment section.

% On the other hand, IWSI-W directly uses the naive estimation $w_i^\prime$ in Eq.~\ref{equ:weight-est} to calculate a weighted training loss:
% \begin{equation}
%     \label{IWSI-W}
%     \mathcal{L}_{W} = \frac{1}{|\mathcal{D}_c|}\sum_{x_i}^{\mathcal{D}_c} Clip( w_i^{\prime},2)\cdot \mathcal{L}(\mathcal{M}_L(x_i))
% \end{equation}
% Noted that $w_i^{\prime}$ ranges from zero to infinity, to stabilize the training process, we clip $w_i^{\prime}$ to $(0, 2]$.

% The two variants of IWSI both have their own advantages. While IWSI-W would not add additional hyperparameters to adjust, IWSI-F has less data to train after further filtering and seems more robust to the potential estimation error of Eq.~\ref{equ:weight-est}. We investigate them both in our experiments.

%% file: sections/experiment.tex
\section{Experiment}
\subsection{Setup}
\subsubsection{Datasets}
We conduct experiments on six datasets across three types of tasks:
\textbf{Arithmetic Reasoning:} gsm8k~\citep{gsm8k} and SVAMP~\citep{SVAMP}. \textbf{Natural Language Inference:} Adversarial NLI subsets~\citep{ANLI}. ANLI-A1 and ANLI-A2 subsets are used. \textbf{Commonsense Reasoning:} OpenBookQA~\citep{OBQA} and StrategyQA~\citep{StrategyQA}.

For all datasets, only the questions are used to self-generate candidate answers. For gsm8k and SVAMP, we keep the original question format, which is the open-ended question. For the other four datasets, we unify the question format to the multiple choice question. The LLM must choose one option as its answer.

To build the valid set, we extract rationales from the original datasets apart from SVAMP, for which we manually write rationales. The size of valid sets varies among different datasets, but none of them exceeds 5\% size of the corresponding training set.
Appendix A provides more details about the split and statistics of all datasets.

\subsubsection{Baselines}
The goal of our experiments is to verify whether incorporating DS weight into the filtering strategy in our proposed approach can help LLMs self-improve. Therefore, given the same base model, we compare IWSI with the fundamental self-improvement framework LMSI~\citep{LMSI}, and some variants that we implement by adopting trendy filtering strategies designed for training LLMs on model-generated data.

LMSI~\citep{LMSI} is the first self-improvement framework that significantly improves LLMs' reasoning ability without any external supervision. The core idea of LMSI is adopting majority voting to select answers that are most likely correct, thus filtering the self-generated data. 

MoT~\citep{MoT} uses entropy to measure the uncertainty of the answers and further filters data. We combine this technique with LMSI and denote it as \textit{Entropy-filter}. 

Self-Alignment~\citep{Self-alignment} shows that LLM self-evaluation could be helpful in filtering strategy. We implement this idea with LMSI and denote it as \textit{Self-filter}. 
% The prompt used for Self-filter is in Appendix~\ref{sec:prompts}.

Works like GENIE~\citep{GENIE} and ReST~\citep{rest} use pre-trained models to evaluate the self-generated samples. Intervened by external supervision, their filtering results are usually more comprehensive and meticulous. Following that, we also implement a variant of LMSI for reference, the \textit{RM-filter}. RM-filter uses a pre-trained reward model to score the generated data, as GENIE~\citep{GENIE} does.~\footnote{https://huggingface.co/OpenAssistant/reward-model-deberta-v3-large-v2}

\input{tables/main}

\subsubsection{Implementation details}
We select Llama3-8B as our base model~\citep{Llama}. 
% In the candidate answers generation stage, we let the base model generate 15 candidates for each question, with temperature $T=1.1$.
For each question, we generate 15 candidates, with temperature $T=1.1$.
All training process is performed on eight RTX-4090 GPUs. The training batch size per device is set to 1 and the gradient accumulation steps is 4. We use LoRA~\citep{lora} to do fine-tuning. We use AdamW~\citep{adamW} optimizer and the learning rate is 3e-4. Few-Shot-CoT prompts are only applied in generating candidate answers and the evaluation stage. CoT examples for each dataset, prompts used for Self-filter, and details about how to derive the answer from output texts are given in Appendix D. The source code and supplementary materials are available at https://github.com/rubickkcibur/IWSI.

\subsection{Main Results}
The main comparison results are shown in Table~\ref{tab:main}. The evaluation metric is accuracy percent and all results are derived by greedy decoding. The top part is the performance of the base model. The middle part are self-improvement baselines and our proposed method IWSI. For reference, we list the performance of RM-filter at the bottom of the table. For fairness, we universally set the filtering percentage $k=80$ for IWSI, Entropy-filter, Self-filter, and RM-filter.

Among self-improvement methods (the middle part), IWSI is the only one that consistently outperforms LMSI, and it also achieves the best in almost all datasets. We further empirically demonstrate that the superiority of IWSI primarily stems from excluding self-generated samples with higher DSE, rather than merely from access to part of the information of the valid set (the mean loss value of valid samples). Details are in Appendix F.
% For different tasks, we found that the improvement bonus is much more significant in arithmetic reasoning tasks than in others. LMSI has a 310\% enhancement in gsm8k and 206\% in SVAMP, compared to the base model. IWSI further outperforms LMSI 34.8\% in gsm8k and 39.3\% in SVAMP. 

% As Table~\ref{tab:main} shows, the effectiveness of self-improvement methods on arithmetic  reasoning task is the most significant. Compared to base model performance, LMSI has a 310\% enhancement in gsm8k and 206\% in SVAMP, and IWSI-F has a 453\% improvement in gsm8k and 327\% in SVAMP. In contrast, the improvement range on other four datasets is apparently smaller. 

% Among self-improvement methods, IWSI consistently outperforms LMSI and almost all variants. It proves that filter out self-generated samples with high DSE does have positive influence on boosting the model performance. The bonus of taking DSE into filtering strategy is also much more significant in arithmetic reasoning than other tasks. Specifically, IWSI-F outperforms LMSI 34.8\% in gam8k and 39.3\% in SVAMP, while only 7.9\% in ANLI-A1 and 6.2\% in StrategyQA. We believe this is because high DSE samples may have a stronger misleading on math questions. 

For IWSI-w, the variant of IWSI that uses DS weight to compute weighted loss other than filtering data, it generally performs worse than IWSI, even though IWSI-w is more compliant with the standard importance weighting formula. The most possible reason is that unlike deep methods like DIW~\citep{DIW}, which uses a deep neural network to learn the weights, our weight estimation (Eq.~\ref{equ:weight-est}) is a pretty naive approach. While it largely reduces computational cost, it also omits the semantic similarity among training samples, potentially compromising efficacy. Therefore, the weighted loss in IWSI-w might make the training process difficult and noisy. In contrast, IWSI only uses the weight as an indicator to rank the samples with respect to DSE, without directly incorporating the weight into the training loss, which makes the overall process more robust. 
%相比于RM-filter，RM在数学上效果更好而在其他数据集上效果不明显

As for the RM-filter, we found that it does not always perform the best among all six datasets, even though it introduces external supervision by using a pre-trained reward model. As Table~\ref{tab:main} shows, after incorporating both the answer correctness and DSE of samples, 
the overall performance of IWSI is comparable to that achieved with external supervision from a pre-trained reward model.

\subsection{Hyperparameter Study}
We investigate the effect of varying the filtering threshold $k$ and corresponding percentile $\sigma_{k\%}$ (in Eq.~\ref{equ:sigma}). Fig.~\ref{fig:k-para} shows the accuracy results on gsm8k, StrategyQA, and ANLI-A1. As the figure shows, either a too-large or too-small $k$ value will make the performance degrade. When $k$ is very large, more samples with high DSE will be kept, thus potentially harming the performance. If the $k$ is pretty small, there will not be sufficient samples kept to support the model training. The optimal $k$ value range varies across different tasks. In general, around 80\% would be an appropriate choice. 

Fig.~\ref{fig:th_bar} shows the varying $k$-percentile $\sigma_{k\%}$ of DS weight. While $\sigma_{k\%}$ of different datasets are similar when $k$ is very small, the difference becomes larger as $k$ increases. This phenomenon suggests that the boundary above which the DSE of samples can be regarded as "high" is relative according to different datasets.
\input{figures/k_para}
\input{figures/th_bar}

\subsection{Valid Set Analysis}
\label{sec:valid-analysis}
The valid set $\mathcal{D}_v$ plays a crucial role in IWSI. 
% Without $\mathcal{D}_v$ we can not calculate DS weight. 
It determines the calculation results of DS weight and subsequently steers the filtering strategy. Therefore, variation in the composition of the valid set can introduce randomness and thus potential instability. In this section, we take the gsm8k dataset as example to discuss the impact of valid set.

We employ the loss value distribution as the analytical tool and, for simplicity, we assume all distributions of different sample sets conform to the normal distribution. For example, the loss value distribution of valid set is denoted as $\mathcal{N}_v(\mu_{v},\sigma_{v}^2)$, where $\mu_v$ and $\sigma_v$ are the \textit{mean} and \textit{standard deviation} respectively.

Fig.~\ref{fig:train_val} shows distributions of the valid set and self-generated samples before and after IWSI. Analogous to our intuition, the distributions differ significantly between valid set samples and self-generated samples before IWSI, and become much closer after IWSI, illustrating the effectiveness of IWSI in handling the distribution shift problem. Furthermore, we provide quantitative analyses and a case study in Appendix E for a better understanding of how the LLM generation was affected by IWSI.

The next question is \textit{would the randomness of valid set composition cause great instability in IWSI}, since $\sigma_v$ is apparently not small enough. The answer is "No" as long as there is an adequate valid set size $N_v$. Theoretically, in Eq.~\ref{equ:weight-est}, it is only the \textit{sample mean}, denoted as $\bar{\mathcal{L}_v}$, that matters.
$\bar{\mathcal{L}_v}$ is also subject to the normal distribution, with its standard deviation inversely proportional to the size $N_v$:
\begin{equation}
    \label{equ:valid_dist}
\begin{aligned}
    \bar{\mathcal{L}_v} &= \frac{\sum_{x_j^v \in \mathcal{D}_v} \mathcal{L}(\mathcal{M}_L(x^v_j))}{N_v} \\
    \bar{\mathcal{L}_v} &\sim \mathcal{N}_{\bar{v}}(\mu_v,(\frac{\sigma_v}{N_v})^2)
\end{aligned}
\end{equation}
Eq.~\ref{equ:valid_dist} implies that increasing $N_v$ can scale down the variance of $\bar{\mathcal{L}_v}$, thus making the estimation more stable. More importantly, \textit{it is completely irrelevant to the size of the training samples}. For instance, in gsm8k, if the valid set size is $100$, the standard deviation of $\bar{\mathcal{L}_v}$ is $\bar{\sigma_v} = \frac{\sigma_v}{100} = 4.1\times 10^{-3}$, which is small enough to mitigate the interference of randomness.

\input{tables/valid_gsm8k}
To empirically investigate the influence of different valid set compositions, we randomly constitute six subsets of the valid set of gsm8k and test IWSI with them. Table~\ref{tab:empirical-gsm8k} shows the results. $N_v$ denotes the valid set size. $\bar{\mathcal{L}_v}$ is the sample mean of different composition. We use \textit{acc} as the metric.

As we can see, the impact of different compositions on the accuracy results is quite minimal. We believe this is primarily attributed to the \textit{double-robustness} of IWSI. First, the DS weight calculation is robust to the valid set composition, since it only uses the sample mean which varies vary little. Furthermore, the filtering strategy is also robust to the DS weight, since the DS weight is used for ranking other than weighting. As a result, samples with extremely high DSE are probably always discarded even if DS weight changes. 

\input{figures/train_valid_gsm8k}

\subsection{Orthogonality Analysis}
\input{figures/othogonality}
In IWSI, two factors are considered in the filtering strategy, the answer correctness (represented by self-consistency) and the sample DSE (represented by DS weight). A natural question is what is the relationship between these two factors. Are they correlated to or independent of each other? To explore this question, we counted the percentage of samples with correct answers (using the ground truth labels) across different DS weight intervals, as Fig~\ref{fig:oth} shows. Along $x$-axis are the selected intervals: $[1, 1.1)$, $[1.1, 1.3)$, $[1.3, 1.5)$, $[1.5, 2)$, and $[2, \infty)$. In each bar, the upper portion (yellow) indicates the ratio of correct answers, while the lower portion (blue) represents the ratio of wrong answers. For all datasets, we observe a general downward trend in the ratio of correct answers, as DS weight increases. The highest ratios of correct answers is found either in the $[1, 1.1)$ interval (for gsm8k and ANLI-A1) or in the $[1.1, 1.3)$ interval (for StrategyQA). However, both correct and wrong answers occupy a portion that can not be ignored in every interval, suggesting a degree of independence between these two factors.

We delve deeper into the relationship between DSE and the answer uncertainty, which is first investigated by MoT~\citep{MoT} regarding its impact on self-improvement. MoT also suggested using entropy to represent answer uncertainty. We briefly introduce the calculation: 
given a certain question $q$, the self-generated candidate answers $[a_1, a_2, \dots, a_m]$, and the most consistent answer $\hat{a}$, uncertainty $u$ is computed in the following steps:
\begin{equation}
    \label{equ:uncertainty}
\begin{aligned}
    &A^{*} = unique(\{a_i\}^m) \\
    &p(a_j^*) = \sum_i^m \vmathbb{1}(a_j^* =a_i) / m \\
    &u = - \sum^{A^*}_{a_j^*} p(a_j^*)\log p(a_j^*)
\end{aligned}
\end{equation}
where $A^{*}=\{a_1^*,a_2^*,\cdots\}$ is the unique answer set. The higher $u$ is, the more uncertain the answer is. In extreme cases, if $u = 0$, all candidate answers are identical, and if each candidate answer has its unique value, $u$ will reach the maximum $\log m$. For convenience, we normalize $u$ with a divisor $\log m$ and we denote the filter threshold as $u^*$.

We draw the probability density function (PDF) of DS weight for various uncertainty thresholds $u^*$. The second row of Fig.~\ref{fig:oth} shows the results. For arithmetic reasoning (gsm8k), as $u^*$ increases, the peak of PDF falls and the PDF curve becomes flatter, indicating a growth in the proportion of samples with high DSE. Conversely, for commonsense reasoning (StrategyQA) and natural language inference (ANLI-A1), the relationship between uncertainty and DSE appears much weaker. The PDF curves are almost identical, with little variation at the peak, suggesting that DSE is nearly orthogonal to the uncertainty. 
%原因分析
% We believe this phenomenon has a lot to do with the question format. The question type of OBQA and ANLI is multiple choice question, and the model can only choose its answer from several candidates. But the type of gsm8k and SVAMP is short answer question and the answer set is infinite, which makes the answer variance more sensitive to the DSE.

\subsection{Perception of DSE}
We conducted a case study on gsm8k to provide an intuitive perception about what a correct but with high DSE sample looks like. We compare the generated answers with the highest and lowest DSE for the same question. We found that cases with the highest DSE are usually notably absurd that we can easily tell them apart from human-written samples. We categorize these samples into 3 types:
\begin{itemize}
    \item \textbf{Redundant samples.} Redundant samples include irrelevant or repeated information in the reasoning thoughts, making it confusing.
    \item \textbf{Jumping samples.} Jumping samples omit essential reasoning steps or even directly give the answer, making it less logically fluent.
    \item \textbf{Spurious samples.} The reasoning steps in a \textit{spurious} sample~\citep{spurious1, spurious2} are logically wrong. They get the correct answer just by coincidence.
\end{itemize}
We give more exact demonstrations in Appendix B.

% \subsection{Task Generalization}
% \subsection{Weak-to-Strong Generalization}

%% file: tables/main.tex
%这个RM-filter放这里乍一看容易误解，可以分开单独做一个table
\begin{table*}
\centering
    \resizebox{0.85\textwidth}{!}{
    \begin{tabular}{lccccccc}
    \toprule
     & \textbf{gsm8k} & \textbf{SVAMP} & \textbf{ANLI-A1} & \textbf{ANLI-A2}  & \textbf{OpenBookQA} & \textbf{StrategyQA} & \textbf{Avg.}\\
    \midrule
    base & 7.0 & 14.7 & 16.4 & 14.6 & 31.8 & 48.3 & 22.1\\
    \midrule
    LMSI & 27.9 & 45.0 & 25.2 & 22.6 & 31.6 & 51.4 & 34.0\\
    Entropy-filter & 22.7 & 56.0 & 25.2 & 22.8 & 33.4 & 51.2 & 35.2\\
    Self-filter & 35.6 & \textbf{62.7} & 22.8 & \textbf{25.6} & 35.0 & 50.4 & 38.7\\
    IWSI-w & 37.0 & 43.3 & 21.8 & 21.8 & 31.8 & 49.2 & 34.2\\
    \textbf{IWSI} & \textbf{37.6} & \textbf{62.7} & \textbf{27.2} & 23.4 & \textbf{37.0} & \textbf{54.6} & \textbf{40.4}\\
    \midrule
    RM-filter & 40.0 & 66.3 & 25.6 & 25.0 & 34.2 & 51.4 & 40.4\\
    \bottomrule
    \end{tabular}
    }
    \caption{Accuracy results on all datasets. Numbers in the table are the accuracy percent. The first part is the performance of the base model. The second part is the performance of three baseline self-improvement methods, our proposed method IWSI, and a variant IWSI-w. As RM-filter uses the external reward model, we list its performance separately at the bottom of the table.}
    \label{tab:main}
\end{table*}

%% file: figures/k_para.tex
\begin{figure}[t]
  \centering
  \includegraphics[width=0.75\columnwidth]{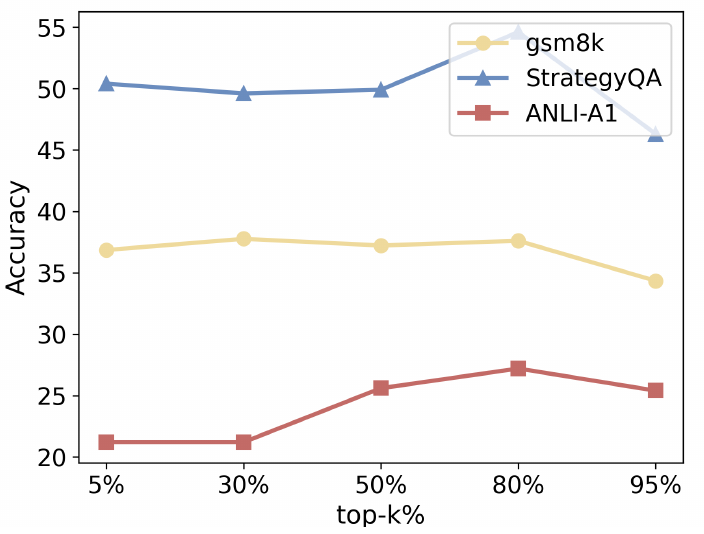}
  \caption{Accuracy results with varying $k$ values.}
  \label{fig:k-para}
\end{figure}

%% file: figures/th_bar.tex
\begin{figure}[t]
  \centering
  \includegraphics[width=0.75\columnwidth]{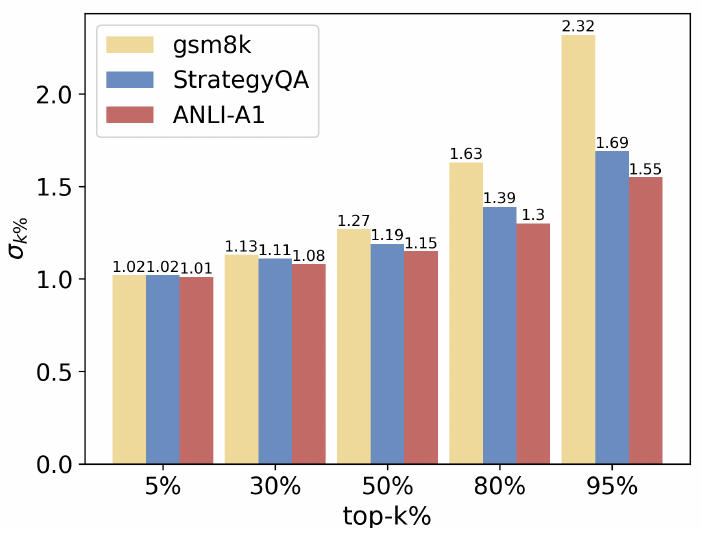}
  \caption{$\sigma_{k\%}$ (in Eq.~\ref{equ:sigma}) with varying $k$ values.}
  \label{fig:th_bar}
\end{figure}

%% file: tables/valid_gsm8k.tex
\begin{table}
    \centering
    \resizebox{0.9\columnwidth}{!}{
    \begin{tabular}{lcccccc}
    \toprule
    \multirow{2}{*}{$N_v$} & \multicolumn{2}{c}{Comp.1} & \multicolumn{2}{c}{Comp.2} & \multicolumn{2}{c}{Comp.3} \\
    \cmidrule(r){2-3} \cmidrule(r){4-5} \cmidrule(r){6-7}
    & $\bar{\mathcal{L}_v}$ & acc & $\bar{\mathcal{L}_v}$ & acc & $\bar{\mathcal{L}_v}$ & acc \\
    \midrule
    50 & 2.053 & 36.77 & 2.083 & 36.54 & 2.091 & 36.47  \\
    100 & 2.077 & 36.69 & 2.086 & 36.47 & 2.054 & 37.00 \\
    \bottomrule
    \end{tabular}
    }
    \caption{Results of different valid sets on gsm8k.}
    \label{tab:empirical-gsm8k}
\end{table}

%% file: figures/train_valid_gsm8k.tex
\begin{figure}[t]
  \centering
  \includegraphics[width=0.8\columnwidth]{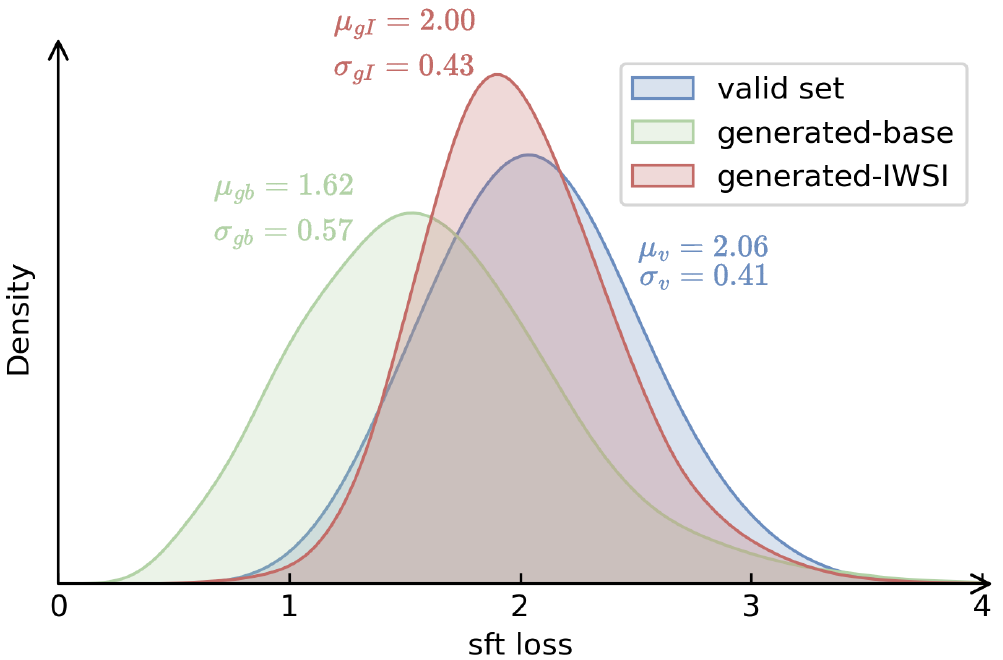}
  \caption{Loss value distributions of the valid set samples, self-generated samples of base model (generated-base), and self-generated samples after IWSI (generated-IWSI), of gsm8k. $\mu$ and $\sigma$ denote the \textit{mean} and \textit{standard deviation}.}
  \label{fig:train_val}
\end{figure}

%% file: figures/othogonality.tex
\begin{figure*}[t]
  \centering
  \label{fig:oth}
  \includegraphics[width=0.95\textwidth]{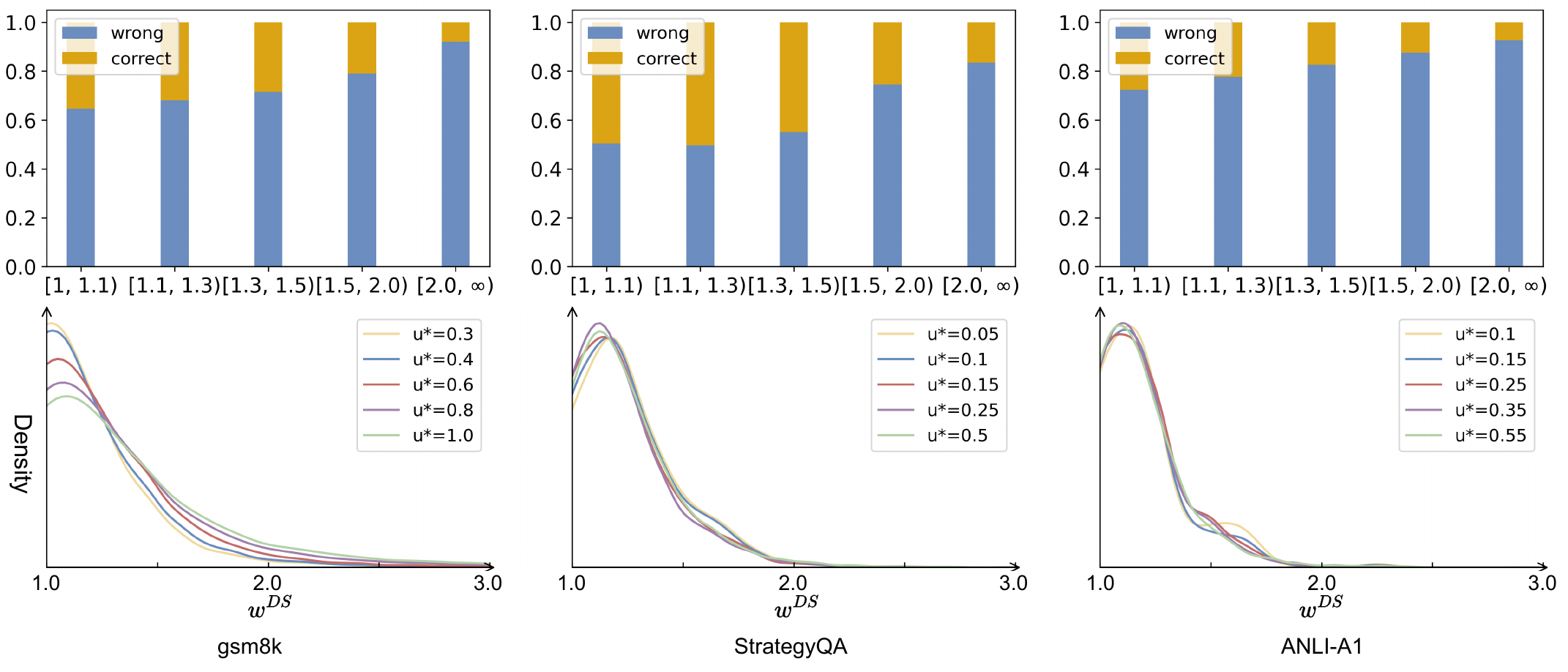}
  \caption{The first row shows relationship between answer correctness and DSE, where along $x$-axis are DSE intervals and the $y$-axis indicates the proportion of correct answers and wrong answers. The second row are the DS weight probability density function curves with varying uncertainty threshold $u^*$ (Eq.~\ref{equ:uncertainty}).}
  \label{fig:oth}
\end{figure*}

%% file: sections/conclusion.tex
\section{Conclusion}
In this paper, we investigate the impact of sample DSE on LLM self-improvement.
% how importance weighting can benefit LLM self-improvement. 
We propose \textit{DS weight} to approximate the DSE inspired by importance weighting methods,
% of self-generated samples, 
and a novel framework IWSI where the filtering strategy comprehensively considers DSE and answer correctness. Empirical results demonstrate that the incorporation of DS weight significantly enhances the effectiveness of LLM self-improvement. Further analysis reveals that DSE is nearly orthogonal to other factors, suggesting a new direction to promote LLM self-improvement for the future work.
% Further analysis discuss the influence of the hyperparameter and the randomness of valid set constitution. It is also found that the relationship between DSE and answer correctness, as well as uncertainty, is orthogonal in a certain degree.

%% file: sections/appendix.tex
\section{Appendix}
\subsection{A Datasets Details}
\label{sec:statistics}
\input{tables/datasets}
Table~\ref{tab:statistics} shows the statistics of six datasets used in our paper. $|\mathcal{D}_q|$, $|\mathcal{D}_v|$, and $|\mathcal{D}_t|$ indicate the sizes of the unsupervised training dataset, valid set, and test set correspondingly. For gsm8k, SVAMP, StrategyQA, and OpenBookQA, we use the whole dataset. The original training set is split into $\mathcal{D}_q$ and $\mathcal{D}_v$. Note that $\mathcal{D}_q$ only contains unlabeled questions. We build subsets for ANLI-A1 and ANLI-A2. $\mathcal{D}_q$, $\mathcal{D}_v$, and $\mathcal{D}_t$ are all extracted from the original training set, valid set, and test set. $\mu_v$ and $\sigma_v$ represent the mean and standard deviation estimated in the valid set.

\subsection{B Correct but with high DSE Examples}
\label{sec:case-study}
\input{tables/samples}
We do a case study to investigate what a correct but with high DSE sample looks like. Among 15 candidates answers to one same question, we pick up the sample with the highest DS weight and the one with the lowest DS weight. Table~\ref{tab:case} are some examples. A1 for Q1 is a jumping sample. It directly give the answer after repeating the question, without any thoughts. A1 for Q2 is a spurious sample. Its reasoning process is totally wrong, but coincidentally it gets to the correct answer. A1 for Q3 and A1 for Q4 are redundant samples. Their statement has too much nonsense, making the reasoning thoughts inconsistent and confusing.

In contrast, the A2 for each question is obviously much more reasonable and coherent. Their DS weights are also very close to 1.

\subsection{C Impacts on the Self-Consuming Loop}
\label{sec:valid-size}
\input{figures/iters}

Despite the attractive application prospects of LLM self-improvement, recent studies have shown that, without external validation and correction, iteratively training on self-generated contents can irreversibly defect the resulting model performance, which is called \textit{model collapse}~\citep{moddel-collapse}. This phenomenon occurs because the tails of the original content distribution disappear, making the distribution more concentrated and consequently decreasing the diversity of generated samples.

Compared to a typical self-consuming loop, IWSI uses the DS weight to filter generated samples, making the sample distribution closer to the valid sample distribution. We would like to investigate its impact on the model collapse procedure. Fig.~\ref{fig:iters} are the accuracy results with self-consuming iterations. We compare IWSI with the \textit{self-filter}. For both of them, on each iteration, we train from scratch with samples generated from the resulting model of the last iteration. As we can see, the performance of IWSI decreases much slower than the self-filter, especially at the beginning, showing a mitigating effect on model collapse. 

However, the results also indicate that IWSI can not stop the model collapse procedure.

\subsection{D More Implementation Details}
\label{sec:prompts}
\input{tables/cot_prompts}
\input{tables/self-prompt}
\input{tables/main_first}
Table~\ref{tab:gsm8k}, Table~\ref{tab:ANLI}, Table~\ref{tab:OBQA}, and Table~\ref{tab:strategyQA} are the Chain-of-Thought examples for corresponding tasks, used in our experiments. Table~\ref{tab:self} is the prompt used to let the LLM self-score its generated answers in the Self-filter baseline. This prompt is adapted from~\citep{Self-alignment}. We do not copy the original one since it is designed for evaluating the instruction following tasks. 

When calculating the accuracy for each task, we need to parse the generated outputs of language models. In our setting, we require the LLM to give answers by following \textit{"The answer is"} format. We treat outputs that do not follow the required format as false. For outputs where more than one \textit{"The answer is"} exist, we found that, although the absolute values of the results may differ depending on whether the first or last occurrence is taken, the comparative results are broadly similar. Therefore, the conclusions drawn from this will not change as a result. The results in Table~\ref{tab:main} are calculated by taking the \textbf{last} occurrence. We also provide the results by taking the \textbf{first} occurrence in Table~\ref{tab:main_first} for reference.

\subsection{E Analysis Between Generated and Valid Samples}
\label{sec:sample_comparison}
\input{tables/valid_generated}
\input{tables/with_valid}
\input{tables/semantics}
\input{tables/lexical}
\input{tables/sytactic}

In the experiment section, we give a general finding that IWSI can steer the loss value distribution of self-generated samples to that of the valid set (Fig.~\ref{fig:train_val}). In this section, we discuss how the LLM generation was influenced after IWSI in greater detail.

First, we provide a case study to investigate \textit{in what aspect the generated samples after IWSI are similar to the valid samples}. We select several questions in the valid set, let the model after IWSI generate answers using greedy decoding, and compare them with the answers in the valid set correspondingly. Table~\ref{tab:valid-gene} shows the cases, where five generated answers are correct (Q1-Q5) and others are wrong (Q6-Q8). For the correct cases, the most common feature between the generated sample and valid sample is the calculation equations, which are almost totally identical and follow the same order. Meanwhile, compared to valid answers, generated answers have more logical connectors, which makes the length of words longer. However, for those cases where the self-generated answers are wrong, there is no apparent pattern. Their mistakes can be caused by logical disordering (Q6), omitting important information (Q7), or twisting the conditions of the original question (Q8).

For further quantitative analysis, we let the LLMs (before and after IWSI) generate answers to the valid questions, and measure the semantic similarity, lexical diversity, and syntactic diversity of these samples. 

Table~\ref{tab:semantic} are the semantic similarity results. For each question in the valid set, we use the SimCSE model~\footnote{https://huggingface.co/princeton-nlp/sup-simcse-bert-base-uncased} to obtain the embedding vectors of the valid answer and generated answer, and compute their cosine similarity. As we can see, the semantic similarity improves after IWSI in general, and the promotion is more significant on the NLI tasks, which may be because the reasoning in the NLI task relies more heavily on semantic information.

We measure the lexical diversity by the Type Token Ratio (TTR) metric, which is calculated as the ratio of unique words in a sentence~\citep{Diversity}. A higher value of TTR indicates more diverse vocabulary usage. Table~\ref{tab:ttr} shows the results. As we can see, the valid answers have higher TTR than model-generated answers among all datasets, showing that human annotations possess more diversity. Interestingly, model-generated answers after IWSI tend to have a slightly lower TTR in arithmetic reasoning tasks and a much higher TTR in NLI tasks and commonsense reasoning tasks. 

We measure the syntactic diversity by calculating the average dependency distance (DD). It refers to the distance between two words in a sentence that are connected by a dependency relation~\citep{Diversity}. Generally, a longer distance indicates a complex syntactic structure. We use the \textit{Spacy} library~\footnote{https://spacy.io/} to calculate DD for each sample. Table~\ref{tab:add} shows the results. The average DD is roughly the same across arithmetic and NLI tasks, but in commonsense reasoning tasks, the average DD gets obvious improvement after IWSI, being closer to that of valid samples.

\subsection{F Training with Valid Samples}
\label{sec:valid-training}
In IWSI, we use annotated valid samples to compute DS weight, which is further used to filter the self-generated samples. A natural question is whether the performance promotion is attributed to nothing but access to the valid set. To verify this, we add the valid samples to the training set and compare IWSI with \textit{Self-filter}. In this setting, both IWSI and the baseline method have the information of the valid set.

As Table~\ref{tab:with-valid} shows, under the setting where the training set contains both valid samples and self-generated samples, IWSI can still maintain superiority over the baseline method. We believe the reason is that, as our motivation suggested, the distribution of self-generated samples filtered merely by answer correctness may shift significantly from the desired distribution, thus misleading the training process. This deficiency exists in the self-generated samples and is independent of whether or not valid set samples are accessible.

%% file: tables/datasets.tex
\begin{table}
    \resizebox{\columnwidth}{!}{
    \begin{tabular}{lccccc}
    \toprule
    Datasets & $|\mathcal{D}_q|$ & $|\mathcal{D}_v|$ & $|\mathcal{D}_t|$ & $\mu_v$ & $\sigma_v$ \\
    \midrule
    gsm8k & 7,099 & 374 & 1,319 & 2.06 & 0.41 \\
    SVAMP & 665 & 35 & 300 & 2.43 & 0.32 \\
    ANLI-A1 & 2,000 & 100 & 500 & 2.49 & 0.31 \\
    ANLI-A2 & 2,000 & 100 & 500 & 2.60 & 0.33 \\
    StrategyQA & 1,580 & 23 & 687 & 3.67 & 0.54 \\
    OpenBookQA & 4,750 & 207 & 500 & 4.41 & 0.48 \\
    \bottomrule
    \end{tabular}
    }
    \caption{Statistics of six datasets used.}
    \label{tab:statistics}
\end{table}

%% file: tables/samples.tex
\begin{table*}
\small
\centering
    % \resizebox{\textwidth}{!}{
    \begin{tabularx}{\textwidth}{m{0.9\textwidth}m{0.1\textwidth}}
    \toprule
    \textbf{Q1:} Natalia sold clips to 48 of her friends in April, and then she sold half as many clips in May. How many clips did Natalia sell altogether in April and May? & $w_i^{DS}$ \\
    \midrule
    \textbf{A1:} 48 friends bought clips in April. Half as many friends bought clips in May. How many clips did Natalia sell altogether in April and May? The answer is 72. & 1.63 \\
    \textbf{A2:} 48 of Natalia’s friends had clips in April. She then sold half as many clips in May. To find $\frac{1}{2}$ of 48, we divide 48 by 2. 48 ÷ 2 = 24. So, Natalia sold 24 clips in May. In total Natalia sold 48 + 24 = 72 clips. The answer is 72. & 1.01 \\
    \midrule

    \textbf{Q2:} Weng earns \$12 an hour for babysitting. Yesterday, she just did 50 minutes of babysitting. How much did she earn? & $w_i^{DS}$ \\
    \midrule
    \textbf{A1:} 12 dollars $\times$ 50 minutes = 600 minutes. 600 minutes = 10 hours. So, Weng earned \$10. The answer is 10. & 1.71 \\
    \textbf{A2:} 50 minutes is 50 ÷ 60 = 5/6 of an hour. So, Weng earns \$12 an hour for babysitting. Yesterday, she just did 50 minutes of babysitting. So, she earns \$12 x 5/6 = \$10. Weng earned \$10 yesterday. The answer is 10. & 1.02 \\
    \midrule

    \textbf{Q3:} Betty is saving money for a new wallet which costs \$100. Betty has only half of the money she needs. Her parents decided to give her \$15 for that purpose, and her grandparents twice as much as her parents. How much more money does Betty need to buy the wallet? & $w_i^{DS}$ \\
    \midrule
    \textbf{A1:} 1. A wallet costs \$100. 2. Betty has only half of the money she needs. 3. Her parents decided to give her \$15 for that purpose. 4. Her grandparents twice as much as her parents. 5. Betty needs to buy a wallet. 6. How much more money does Betty need to buy the wallet? 1. Betty has \$50. Her wallet costs \$100. 1. \$100 - \$50 = \$50. 2. Betty needs \$50 more to buy the wallet. 3. Betty’s parents gave her \$15. 4. Betty’s grandparents gave her \$30 [twice as much as her parents did, \$15 + \$15 = \$30]. 2. Betty has \$50. Her parents gave her \$15. Her grandparents gave her \$30. 2. \$50 + \$15 + \$30 = \$95. 3. \$100 - \$95 = \$5. 4. Betty needs \$5 more to buy the wallet. 5. Betty needs to buy a wallet. 6. How much more money does Betty need to buy the wallet? 1. Betty has \$50. Her wallet costs \$100. 2. Betty needs \$50 more to buy the wallet. 3. Betty’s parents gave her \$15. 4. Betty’s grandparents gave her \$30. The answer is 5. & 2.13 \\
    \textbf{A2:} 100 is the price of the new wallet. So Betty needs \$100 to buy the wallet. Betty has only half of the money she needs. Her parents decided to give her \$15. Her grandparents gave her twice as much as her parents. That means they gave her 2 × 15 = 30 more dollars. So, Betty’s grandparents gave her a total of 30 + 15 = 45 new dollars. That means she now has 45 + 50 = 95 new dollars. So she needs 100 - 95 = 5 new dollars more in total. The answer is 5. & 1.19 \\
    \midrule

    \textbf{Q4:} Albert is wondering how much pizza he can eat in one day. He buys 2 large pizzas and 2 small pizzas. A large pizza has 16 slices and a small pizza has 8 slices. If he eats it all, how many pieces does he eat that day? & $w_i^{DS}$ \\
    \midrule
    \textbf{A1:} 2 large pizzas and 2 small pizzas have been bought. Let’s find out how many slices of pizza are there in total. The number of slices of pizza in total will be the number of pieces of pizza he can eat that day. Let’s start. Each large pizza has 16 slices. So, 2 large pizzas have 2 × 16 = 32 slices. Each small pizza has 8 slices. So, 2 small pizzas have 2 × 8 = 16 slices. In total, 32 + 16 = 48 slices of pizza have been bought. Albert has eaten it all. So, he has eaten 48 pieces of pizza that day. The answer is 48. & 1.55 \\
    \textbf{A2:} 2 large pizzas have 2 X 16 = 32 slices. 2 small pizzas have 2 X 8 = 16 slices. 32 + 16 = 48 slices in total. The answer is 48. & 1.06 \\
    \bottomrule
    \end{tabularx}
    % }
    \caption{Case study on gsm8k}
    \label{tab:case}
\end{table*}

%% file: figures/iters.tex
\begin{figure}[t]
  \centering
  \includegraphics[width=0.8\columnwidth]{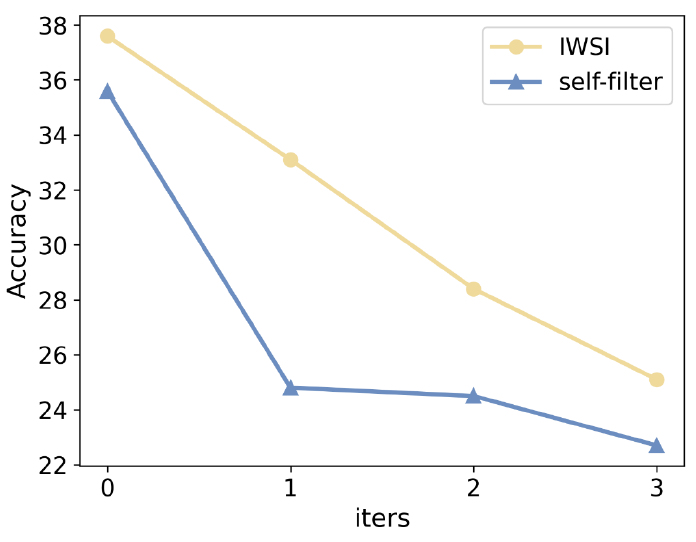}
  \caption{Accuracy results with self-consuming iterations.}
  \label{fig:iters}
\end{figure}

%% file: tables/cot_prompts.tex
\begin{table*}
\small
\centering
    \begin{tabularx}{\textwidth}{p{0.95\textwidth}}
    \toprule
    \textbf{Q:} There are 15 trees in the grove. Grove workers will plant trees in the grove today. After they are done, there will be 21 trees. How many trees did the grove workers plant today? \\
    \textbf{A:} We start with 15 trees. Later we have 21 trees. The difference must be the number of trees they planted. So, they must have planted 21 - 15 = 6 trees. The answer is 6.\\
    \textbf{Q:} If there are 3 cars in the parking lot and 2 more cars arrive, how many cars are in the parking lot? \\
    \textbf{A:} There are 3 cars in the parking lot already. 2 more arrive. Now there are 3 + 2 = 5 cars. The answer is 5. \\
    \textbf{Q:} Leah had 32 chocolates and her sister had 42. If they ate 35, how many pieces do they have left in total? \\
    \textbf{A:} Leah had 32 chocolates and Leah’s sister had 42. That means there were originally 32 + 42 = 74 chocolates. 35 have been eaten. So in total they still have 74 - 35 = 39 chocolates. The answer is 39.\\
    \bottomrule
    \end{tabularx}
    \caption{CoT prompts for gsm8k and SVAMP, extracted from~\citep{CoT}}
    \label{tab:gsm8k}
\end{table*}

\begin{table*}
\small
\centering
    \begin{tabularx}{\textwidth}{p{0.95\textwidth}}
    \toprule
    \textbf{Q:} "Conceptually cream skimming has two basic dimensions - product and geography." Based on this premise, can we conclude the hypothesis "Product and geography are what make cream skimming work." is true? Answer Choices are: A)yes; B)no; C)impossible to tell. \\
    \textbf{A:} Based on "cream skimming has two basic dimensions" we can not infer that these two dimensions are what make cream skimming work. It is not possible to tell. The answer is (C).\\
    \textbf{Q:} "One of our member will carry out your instructions minutely." Based on this premise, can we conclude the hypothesis "A member of my team will execute your orders with immense precision." is true? Answer Choices are: A)yes; B)no; C)impossible to tell. \\
    \textbf{A:} "one of" means the same as "a member of", "carry out" means the same as "execute", and "minutely" means the same as "immense precision". So we can say yes. The answer is (A) \\
    \textbf{Q:} "Fun for adults and children." Based on this premise, can we conclude the hypothesis "Fun for only children." is true? Answer Choices are: A)yes; B)no; C)impossible to tell.' \\
    \textbf{A:} "adults and children" contradicts "only children". So we can not conclude the hypothesis. The answer is (B).\\
    \bottomrule
    \end{tabularx}
    \caption{CoT prompts for ANLI-A1 and ANLI-A2, adapted from~\citep{ANLI-prompts}}
    \label{tab:ANLI}
\end{table*}

\begin{table*}
\small
\centering
    \begin{tabularx}{\textwidth}{p{0.95\textwidth}}
    \toprule
    \textbf{Q:} Poison causes harm to which of the following? Answer Choices are: A)a Tree; B)a robot; C)a house; D)a car. \\
    \textbf{A:} Poison will harm living things, only a tree is a living thing. The answer is (A).\\
    \textbf{Q:} As you look deeper into a Marbel you can see? Answer Choices are: A)the future; B)minut defects; C)colors; D)the other side. \\
    \textbf{A:} Marbel is not transparent, so you can not see the other side. Marbel does not necessarily have multiple colors. You will see minut defects. The answer is (B) \\
    \textbf{Q:} When food is reduced in the stomach? Answer Choices are: A)the mind needs time to digest; B)take a second to digest what I said; C)nutrients are being deconstructed; D)reader’s digest is a body of works. \\
    \textbf{A:} The food is being deconstructed in the stomach during digestion. The answer is (C).\\
    \bottomrule
    \end{tabularx}
    \caption{CoT prompts for OpenBookQA, adapted from~\citep{ANLI-prompts}}
    \label{tab:OBQA}
\end{table*}

\begin{table*}
\small
\centering
    \begin{tabularx}{\textwidth}{p{0.95\textwidth}}
    \toprule
    \textbf{Q:} Do hamsters provide food for any animals? Answer Choices are: A)yes; B)no. \\
    \textbf{A:} Hamsters are prey animals. Prey are food for predators. Thus, hamsters provide food for some animals. The answer is (A).\\
    \textbf{Q:} Could Brooke Shields succeed at University of Pennsylvania? Answer Choices are: A)yes; B)no. \\
    \textbf{A:} Brooke Shields went to Princeton University. Princeton University is about as academically rigorous as the University of Pennsylvania. Thus, Brooke Shields could also succeed at the University of Pennsylvania. The answer is (A). \\
    \textbf{Q:} Yes or no: Hydrogen’s atomic number squared exceeds number of Spice Girls? Answer Choices are: A)yes; B)no. \\
    \textbf{A:} Hydrogen has an atomic number of 1. 1 squared is 1. There are 5 Spice Girls. Thus, Hydrogen’s atomic number squared is less than 5. The answer is (B).\\
    \bottomrule
    \end{tabularx}
    \caption{CoT prompts for StrategyQA, adapted from~\citep{CoT}}
    \label{tab:strategyQA}
\end{table*}

%% file: tables/self-prompt.tex
\begin{table*}
\small
\centering
    \begin{tabularx}{\textwidth}{p{0.95\textwidth}}
    \toprule
    Below is a question and a candidate answer. Evaluate whether or not the answer is a good example. A good answer should be complete, clear, and comprehensive. The answer sentence should be well organized without missing or irrelevant information. Use a number between 0 and 10 to represent the rating of the candidate answer. 10 means the best and 0 means the worst. Please follow the format 'Score: \textless rating\textgreater'. \\ Here are the question and candidate answer: \\
    \textit{\textless Question\textgreater} \\
    \textit{\textless Generated Answers\textgreater} \\
    Score: \\
    \bottomrule
    \end{tabularx}
    \caption{Self-score prompts used in Self-filter, adapted from~\citep{Self-alignment}}
    \label{tab:self}
\end{table*}

%% file: tables/main_first.tex
\begin{table*}
\centering
    \resizebox{0.8\textwidth}{!}{
    \begin{tabular}{lccccccc}
    \toprule
     & \textbf{gsm8k} & \textbf{SVAMP} & \textbf{ANLI-A1} & \textbf{ANLI-A2}  & \textbf{OpenBookQA} & \textbf{StrategyQA} & \textbf{Avg.}\\
    \midrule
    LMSI & 47.0 & 68.7 & 37.2 & 34.6 & 64.5 & 71.4 & 53.9\\
    Entropy-filter & 48.8 & 69.7 & 40.0 & 35.0 & 63.8 & 70.2 & 54.6\\
    Self-filter & 50.1 & \textbf{70.6} & \textbf{41.6} & 35.6 & \textbf{64.9} & 71.2 & 55.7\\
    IWSI-w & 47.6 & 69.0 & 38.6 & \textbf{37.0} & 63.8 & 71.4 & 54.6\\
    \textbf{IWSI} & \textbf{51.5} & 70.3 & \textbf{41.6} & \textbf{37.0} & \textbf{64.9} & \textbf{72.2} & \textbf{56.3}\\
    \midrule
    RM-filter & 49.9 & 67.0 & 41.6 & 38.8 & 64.8 & 70.8 & 55.5\\
    \bottomrule
    \end{tabular}
    }
    \caption{Accuracy results evaluated by taking the first occurrence.}
    \label{tab:main_first}
\end{table*}

%% file: tables/valid_generated.tex
\begin{table*}
\small
\centering
    % \resizebox{\textwidth}{!}{
    \begin{tabularx}{\textwidth}{p{0.95\textwidth}}
    \toprule
    
    \textbf{Q1:} Alex has 4 new shirts. Joe has 3 more new shirts than Alex. If Ben has eight more new shirts than Joe, how many new shirts does Ben have?  \\
    \midrule
    \textbf{V:} Joe has 4 + 3 = 7 new shirts. Ben has 8 + 7 = 15 new shirts. The answer is 15.  \\
    \textbf{G:} Alex has 4 new shirts.  Joe has 3 more new shirts than Alex.  So, Joe has 4 + 3 = 7 new shirts.  Ben has 8 more new shirts than Joe.  So, Ben has 7 + 8 = 15 new shirts.  The answer is 15. \\
    
    \midrule
    \textbf{Q2:} John buys 3 barbells and gives \$850 and gets \$40 in change.  How much did each barbell cost?  \\
    \midrule
    \textbf{V:} The total price of the barbells was 850 - 40 = \$810. So each one cost 810 / 3 = \$270. The answer is 270.  \\
    \textbf{G:} John buys 3 barbells and gives \$850.  He gets \$40 in change.  That means he paid \$850 - \$40 = \$810.  The barbells cost \$810.  There are 3 barbells.  So each barbell must have cost \$810 ÷ 3 = \$270.  The answer is 270. \\
    
    \midrule
    \textbf{Q3:} Angus has \$20 worth of tokens less than Elsa has. If Elsa has 60 tokens and each token is worth \$4, how many tokens does Angus have?  \\
    \midrule
    \textbf{V:} If each token is worth \$4, then \$20 worth of tokens is \$20 / \$4 = 5 tokens. Angus has 5 tokens less than Elsa who has 60 so Angus has 60 - 5 = 55 tokens. The answer is 55  \\
    \textbf{G:} 1st, we find out how many tokens Angus has. We know that Angus has \$20 worth of tokens less than Elsa has. We also know that Elsa has 60 tokens and each token is worth \$4. So, Elsa has 60 x \$4 = \$240 worth of tokens. Angus has \$20 worth of tokens less than Elsa has. So, Angus has \$240 - \$20 = \$220 worth of tokens. We know that each token is worth \$4. So, Angus must have 220 / \$4 = 55 tokens. The answer is 55. \\
    
    \midrule
    \textbf{Q4:} A group of science students went on a field trip. They took 9 vans and 10 buses. There were 8 people in each van and 27 people on each bus. How many people went on the field trip?  \\
    \midrule
    \textbf{V:} The vans held 9 vans * 8 people = 72 people. The buses held 10 buses * 27 people = 270 people. The total number of people on the field trip is 72 + 270 = 342 people. The answer is 342  \\
    \textbf{G:} 9 vans and 10 buses were taken on the field trip. There were 8 people in each van and 27 people on each bus. So in total there were 9 x 8 = 72 people in the vans and 10 x 27 = 270 people on the buses. In total there were 72 + 270 = 342 people. The answer is 342. \\
    
    \midrule
    \textbf{Q5:} Tara has a shoebox that is 4 inches tall and 6 inches wide. She puts a square block inside that is 4 inches per side. How many square inches of the box are left uncovered?  \\
    \midrule
    \textbf{V:} The shoebox is 24 square inches because 4 x 6 = 24. The block is 16 square inches because 4 x 4 = 16. There are 8 square inches uncovered because 24 - 16 = 8. The answer is 8  \\
    \textbf{G:} 4 inches tall and 6 inches wide. That means the box is 4 inches by 6 inches. That is 4 x 6 = 24 square inches. The block is 4 inches per side. That means it is 4 x 4 = 16 square inches. So, the box has 24 - 16 = 8 square inches left uncovered. The answer is 8. \\

    \midrule
    \textbf{\textcolor{red}{Q6}:} Martha has 19 butterflies in her collection. She has twice as many blue butterflies as yellow butterflies. The rest of her butterflies are black. If Martha has 6 blue butterflies, how many black butterflies does she have?  \\
    \midrule
    \textbf{\textcolor{red}{V}:} Martha has 6 / 2 = 3 yellow butterflies. Martha has 6 + 3 = 9 blue and yellow butterflies. Martha has 19 - 9 = 10 black butterflies. The answer is 10.  \\
    \textbf{\textcolor{red}{G}:} 19 butterflies are in Martha’s collection. She has twice as many blue butterflies as yellow butterflies. The rest of her butterflies are black. 6 blue butterflies are in her collection. So, there must be 6 blue butterflies and 6 yellow butterflies. That means there are 6 + 6 = 12 yellow butterflies. The rest of her butterflies are black. So, there must be 19 - 12 = 7 black butterflies. The answer is 7. \\

    \midrule
    \textbf{\textcolor{red}{Q7}:} Tonya spent \$90.00 on art supplies.  4 canvases cost \$40.00 and a mixed set of paints cost 1/2 that much. She also spent \$15.00 on an easel, and the rest of the money on paintbrushes. How much money did she spend on paintbrushes?  \\
    \midrule
    \textbf{\textcolor{red}{V}:} The canvases cost \$40.00 and she spent half that amount on paint so she spent 40 / 2 = \$20.00 on paint. She spent \$40.00 on canvases, \$20.00 on paint and \$15.00 on an easel so she spent 40 + 20 + 15 = \$75.00. She spent \$90.00 total and her other purchases totaled \$75.00 so she spent 90 - 75 = \$15.00 on paint brushes. The answer is 15  \\
    \textbf{\textcolor{red}{G}:} 4 canvases cost \$40.00 and a mixed set of paints cost 1/2 that much. So, the mixed set of paints cost \$40.00 / 2 = \$20.00. The rest of the money she spent on paintbrushes. So, she must have spent \$90.00 - \$40.00 - \$20.00 = \$30.00 on paintbrushes. The answer is 30. \\

    \midrule
    \textbf{\textcolor{red}{Q8}:} At a laundromat, it costs \$4 for a washer and a quarter for every 10 minutes in the dryer. Samantha does 2 loads of laundry in the wash and then divides all the washed clothes equally into 3 separate dryers. If she lets the dryers run for 40 minutes each, how much does she spend altogether?  \\
    \midrule
    \textbf{\textcolor{red}{V}:} Samantha spends \$4 x 2 = \$8 on washing. Each dryer costs 25 x 4 = 100 cents = \$1. Three dryers cost 1 x 3 = \$3. Altogether, she spends 8 + 3 = \$11. The answer is 11.  \\
    \textbf{\textcolor{red}{G}:} 2 loads of laundry in the wash cost \$4 each. So, she spends \$4 x 2 = \$8 on the wash. The dryers run for 40 minutes each. So, she spends 3 x 40 = 120 minutes in the dryers. Each minute costs a quarter. So, she spends 120 x 0.25 = \$30 in the dryers. In total she spends \$8 + \$30 = \$38. The answer is 38. \\
    \bottomrule
    \end{tabularx}
    % }
    \caption{Case study about answers comparison between valid and self-generated samples. \textbf{V} and \textbf{G} indicate whether the answer is from the \textbf{V}alid set or is self-\textbf{G}enerated.}
    \label{tab:valid-gene}
\end{table*}

%% file: tables/with_valid.tex
\begin{table}[h]
    \resizebox{\columnwidth}{!}{
    \begin{tabular}{lcccccc}
    \toprule
     & \textbf{gsm8k} & \textbf{SVAMP} & \textbf{ANLI1} & \textbf{ANLI2}  & \textbf{OBQA} & \textbf{StrategyQA} \\
    \midrule
    Self-filter & 39.3 & 60.0 & 36.4 & 36.0 & 43.0 & 48.9 \\
    IWSI & 38.9 & 60.3 & 38.2 & 38.6 & 45.4 & 54.0 \\
    \bottomrule
    \end{tabular}
    }
    \caption{Accuracy results with the valid samples added into the training set.}
    \label{tab:with-valid}
\end{table}

%% file: tables/semantics.tex
\begin{table}[h]
    \resizebox{\columnwidth}{!}{
    \begin{tabular}{lcccccc}
    \toprule
     & \textbf{gsm8k} & \textbf{SVAMP} & \textbf{ANLI1} & \textbf{ANLI2}  & \textbf{OBQA} & \textbf{StrategyQA} \\
    \midrule
    base & 95.08 & 97.65 & 92.08 & 92.91 & 93.08 & 80.22 \\
    IWSI & 94.76 & 98.22 & 94.54 & 94.99 & 93.70 & 81.56 \\
    \bottomrule
    \end{tabular}
    }
    \caption{The cosine semantic similarity between the model-generated samples and the validation set samples.}
    \label{tab:semantic}
\end{table}

%% file: tables/lexical.tex
\begin{table}[h]
    \resizebox{\columnwidth}{!}{
    \begin{tabular}{lcccccc}
    \toprule
     & \textbf{gsm8k} & \textbf{SVAMP} & \textbf{ANLI1} & \textbf{ANLI2}  & \textbf{OBQA} & \textbf{StrategyQA} \\
    \midrule
    valid & 46.69 & 48.03 & 66.30 & 70.03 & 87.76 & 77.23 \\
    base & 40.93 & 46.97 & 51.22 & 52.35 & 80.78 & 46.97 \\
    IWSI & 39.13 & 46.89 & 63.65 & 56.78 & 82.50 & 66.65 \\
    \bottomrule
    \end{tabular}
    }
    \caption{The Type Token Ratio of valid samples and generated samples before and after IWSI}
    \label{tab:ttr}
\end{table}

%% file: tables/sytactic.tex
\begin{table}[h]
    \resizebox{\columnwidth}{!}{
    \begin{tabular}{lcccccc}
    \toprule
     & \textbf{gsm8k} & \textbf{SVAMP} & \textbf{ANLI1} & \textbf{ANLI2}  & \textbf{OBQA} & \textbf{StrategyQA} \\
    \midrule
    valid & 2.68 & 2.47 & 3.01 & 3.00 & 2.39 & 2.34 \\
    base & 2.70 & 2.47 & 3.12 & 2.96 & 2.21 & 2.21 \\
    IWSI & 2.74 & 2.41 & 3.03 & 2.95 & 2.41 & 2.53 \\
    \bottomrule
    \end{tabular}
    }
    \caption{The average dependency distance of valid samples and generated samples before and after IWSI}
    \label{tab:add}
\end{table}